\documentclass[journal]{IEEEtran}
\usepackage{amsmath,amsfonts}
\usepackage{algorithmic}
\usepackage{algorithm}
\usepackage{array}
\usepackage[caption=false,font=normalsize,labelfont=sf,textfont=sf]{subfig}
\usepackage{textcomp}
\usepackage{stfloats}
\usepackage{url}
\usepackage{verbatim}
\usepackage{graphicx}
\usepackage{cite}
\usepackage{booktabs}
 \usepackage{color} 
\usepackage{makecell}
\usepackage{stfloats}
\usepackage[section]{placeins}
\usepackage{amssymb}
\usepackage{indentfirst}
\usepackage{subfig}
\DeclareUnicodeCharacter{E011}{\ensuremath{-}}
\usepackage[marginal]{footmisc}
\usepackage{multirow}
\usepackage{hyperref}

\begin{document}

\title{Imperceptible Face Forgery Attack via Adversarial Semantic Mask}

\author{Decheng~Liu,~Qixuan~Su,~Chunlei~Peng,~\IEEEmembership{Member, IEEE}~,~Nannan~Wang,~\IEEEmembership{Senior Member, IEEE}~and Xinbo Gao, \IEEEmembership{Fellow, IEEE}
\noindent
\thanks{D. Liu, Q. Su, C. Peng  are with the State Key Laboratory of Integrated Services Networks, School of Cyber Engineering, Xidian University, Xi’an 710071, Shaanxi, P. R. China and with Shanghai Key Laboratory of Computer Software Evaluating and Testing, Shanghai 201112, P. R. China (e-mail: dchliu@xidian.edu.cn; clawersu@gmail.com; clpeng@xidian.edu.cn).\\
N. Wang is with the State Key Laboratory of Integrated Services Networks, School of Telecommunications Engineering, Xidian University, Xi’an 710071, Shaanxi, P. R. China (e-mail: nnwang@xidian.edu.cn).\\
X. Gao is with the Chongqing Key Laboratory of Image Cognition, Chongqing University of Posts and Telecommunications, Chongqing 400065, P. R. China.(e-mail: gaoxb@cqupt.edu.cn).}
}

\markboth{Journal of \LaTeX\ Class Files,~Vol.~14, No.~8, June~2024}%
{Shell \MakeLowercase{\textit{et al.}}: A Sample Article Using IEEEtran.cls for IEEE Journals}


\maketitle

\begin{abstract}
With the great development of generative model techniques, face forgery detection draws more and more attention in the related field. 
Researchers find that existing face forgery models are still vulnerable to adversarial examples with generated pixel perturbations in the global image. 
These generated adversarial samples still can’t achieve satisfactory performance because of the high detectability.
To address these problems, we propose an Adversarial Semantic Mask Attack framework (ASMA) which can generate adversarial examples with good transferability and invisibility.
Specifically, we propose a novel adversarial semantic mask generative model, which can constrain generated perturbations in local semantic regions for good stealthiness.
The designed adaptive semantic mask selection strategy can effectively leverage the class activation values of different semantic regions, and further ensure better attack transferability and stealthiness.
Extensive experiments on the public face forgery dataset prove the proposed method achieves superior performance compared with several representative adversarial attack methods.
The code is publicly available at \href{https://github.com/clawerO-O/ASMA}{https://github.com/clawerO-O/ASMA}.
\end{abstract}

\begin{IEEEkeywords}
Forgery detection, face forgery, adversarial learning, adversarial attack.
\end{IEEEkeywords}

\section{Introduction}

\IEEEPARstart{I}{t} has been found that deep learning-based AI systems are susceptible to being fooled by small well-designed perturbations that make significantly incorrect and high-confidence predictions. 
With the development of image forgery techniques, forgery detection models are under increasing threat.
Researchers often generate adversarial images by attacking algorithms to make detection models misclassify images.
Adversarial images have a great impact on the optimization of forgery detection models in addition to their negative aspects in terms of security and privacy.
The robustness and detection accuracy of the models trained with adversarial samples are improved to varying degrees.
Therefore, the study of attack algorithms for generating adversarial examples has significance and application value in promoting the development of face forgery detection models.
Existing attack algorithms mainly add the generated adversarial noise to the original image, which makes the deep network inference abnormality. 
Because of the large area of adversarial noise, the generated adversarial examples have poor stealth, making them easy for human eyes to recognize.

Thus, it is necessary to explore a stealthy adversarial attack algorithm capable of deceiving both human eyes and forgery detector machines.
However, most existing adversarial attack methods generate adversarial samples by adding adversarial perturbations to the whole face area, without considering the diverse properties of different semantic face regions.
The former methods always generate redundant adversarial noises, which means the adversarial generation model adds too much adversarial noise in the regions irrelevant to the model's decision.
Thus, \emph{it is still a challenge to enhance the attacking ability while generating high-quality faces and maintaining visual stealthiness for face forgery detection models.}
To improve the repeatability and stealthiness of the generated adversarial examples, we propose an adversarial semantic mask attack algorithm for face forgery detection tasks.
We leverage the class activation mapping and face semantic parsing module to locate the key semantic regions adaptively.
Then, we constrain the adversarial attack noise region and enhance the stealthiness of the adversarial sample by adding adversarial noise to the key part of the face image. 
We conducted experiments on the Deepfake Detection Challenge (DFDC) dataset~\cite{dolhansky2020}. 
The effectiveness of the proposed method has been proven by compared with diverse representative adversarial attack algorithms.

The main contributions of our paper can be summarized as follows:

\begin{enumerate}
    \item We explore a novel adversarial semantic mask generation pipeline attacking face forgery detection, which can constrain generated perturbations in local semantic regions for good stealthiness.
    
    \item We further propose the adaptive semantic mask selection strategy, which leverages the class activation mapping to select more suitable adversarial semantic mask regions and aims to maintain low perceptibility in real applications.

    \item Experimental results on public large-scale DFDC dataset illustrate the superior performance of the proposed ASMA compared with representative adversarial attack algorithms.
    The code is publicly available at \href{https://github.com/clawerO-O/ASMA}{https://github.com/clawerO-O/ASMA}.
    
\end{enumerate}

\section{Related Work}
\subsection{Forgery Detection Methods}
\label{A.Fake face detection}
Over the past decade, with the emergence and continued maturation of deep learning techniques, forgery techniques, especially image forgery, have had a significant impact on people's lives.
With the development of adversarial generative models, high-quality forged images are becoming increasingly difficult for humans to recognize correctly.
Regarding the potential malice from academics, the researchers attempted to detect whether the images had been tampered with to mitigate the danger, which was seen as a binary classification problem.
Existing forgery detection techniques are carried out in two main areas, the spatial domain and the frequency domain.
In the image domain, some works~\cite{bai2023aunet, xiao2023mcs} have utilized the approach of extracting information about the content features of an image to detect unusual noise.
These methods only utilize the information from the spatial domain, which generally overfits the classification boundary.
In the frequency domain, some works~\cite{miao2023f, li2021discriminative} leverage the difference in the frequency domain between real and fake images for forgery detection.
These methods make forgery detection more reliable, but the detection of images takes a longer time.
Also, researchers explore more accurate and efficient detection methods by utilizing both the image domain and frequency domain~\cite{Liu2023HFC}.
However, limited works focus on exploring the adversarial attacking samples for face forgery methods.
\subsection{Adversarial Attack Methods}
Adversarial attack examples aim to mislead deep learning models and transfer across different target models.
Generally, given a well-trained network, the goal of the adversarial attack is to generate adversarial examples that make the network predict wrongly.

\textbf{Adversarial attack noise}.
\emph{Gradient-based attack}.
To mislead the pre-trained model for classification detection, the strength of the attack needs to be increased and the image changed enough to be discerned by the human eye.
The classic case of noise-based adversarial example generation is an experiment conducted by Goodfellow et al.~\cite{goodfellow2014explaining}, which proves that the recognition results of the model can be misled by adding a small amount of perturbation to the image. 
These attack algorithms are usually single-step or multi-step attack methods that calculate the perturbations based on the gradient of the adversarial loss.
Several gradient-based methods have been proposed for adversarial attacks, including the Basic Iterative Method (BIM) and projected gradient descent (PGD).
In addition, a method called DeepFool ensures that the distance is minimized throughout the iteration by calculating the gap between the adversarial sample and the original sample, which minimizes the generated perturbations.
Although adversarial examples can be generated in this way to confuse the model, as the attack intensity increases, the human eye can observe the difference between the source image and the adversarial image.
\emph{Optimization-based attack}.
Like the model training process, researchers have treated the process of generating adversarial samples as a task, taking as a goal to be able to perturb the forgery detection model, and setting up an optimizer such that the adversarial samples are continually tuned to come closest to the model's decision boundaries.
In these methods, the attack is made somewhat model-based by excluding some pixel points that have little effect on the model classification.
These optimization-based methods ensure that the noise range of the antagonistic samples is small, but they also lead to larger time-consuming and less transferability.

\textbf{Adversarial attack on face analysis}.
With the development of forgery technology, forgery detection models are increasingly replacing human eyes as the primary way of forgery detection.
It has also been argued that adversarial image generation can be achieved by substituting face regions, mainly in the form of patches~\cite{komkov2021advhat}. 
Some works~\cite{szegedy2013intriguing, yang2021towards} propose makeup by models extracting the make-up features of the target face image to generate specific noise to be added to the face.
Zihao Xiao et al~\cite{xiao2021improving} extend the proposed GenAP methods to other tasks, e.g., image classification via adversarial patches in the query-free black-box setting.
For example, Yang Hou et al~\cite{hou2023evading} propose a method that can evade forgery detectors by minimizing the statistical differences between natural and fake images.
In this paper, we propose an adversarial semantic mask attack algorithm that can mislead the forgery detection model while ensuring good stealthiness.

\section{Proposed Approach}
To generate images with better mobility and stealthiness, our work leverages the adaptive semantic mask selection strategy and face semantic parsing module to enhance image stealthiness and attack transferability. 
In this section, we start by introducing the motivation for the algorithm. 
The details of the proposed framework are presented. 

\subsection{Motivation}
Existing forgery detection attack methods tend to globally modify the original face through generating adversarial noises, without considering the diverse properties of different face semantic regions. 
This kind of adversarial attack method may be effective when attacking non-face data.
However, considering the specific and rich semantic structure in face images, it is not suitable to directly bring traditional adversarial attack algorithms to the forgery detection attacking task.
This is because this kind of adversarial noise in the global area easily generates unnatural visual artifacts, which can result in high detectability.
Existing related methods only focus on the whole face area and do not take into account the specific properties of different semantic regions, so the generated adversarial samples may add redundant disturbance in the region unrelated to the model decision. 
The adversarial samples have obvious adversarial artifacts, which are easy to be perceived by the human eye, resulting in low stealthiness. 
Considering the specific properties of different face semantic regions, we propose the novel adversarial semantic mask attacking face forgery detection method to maintain both high transferability and low detectability.

\begin{figure*}[th]
    \centering
    \includegraphics[width=0.9\textwidth]{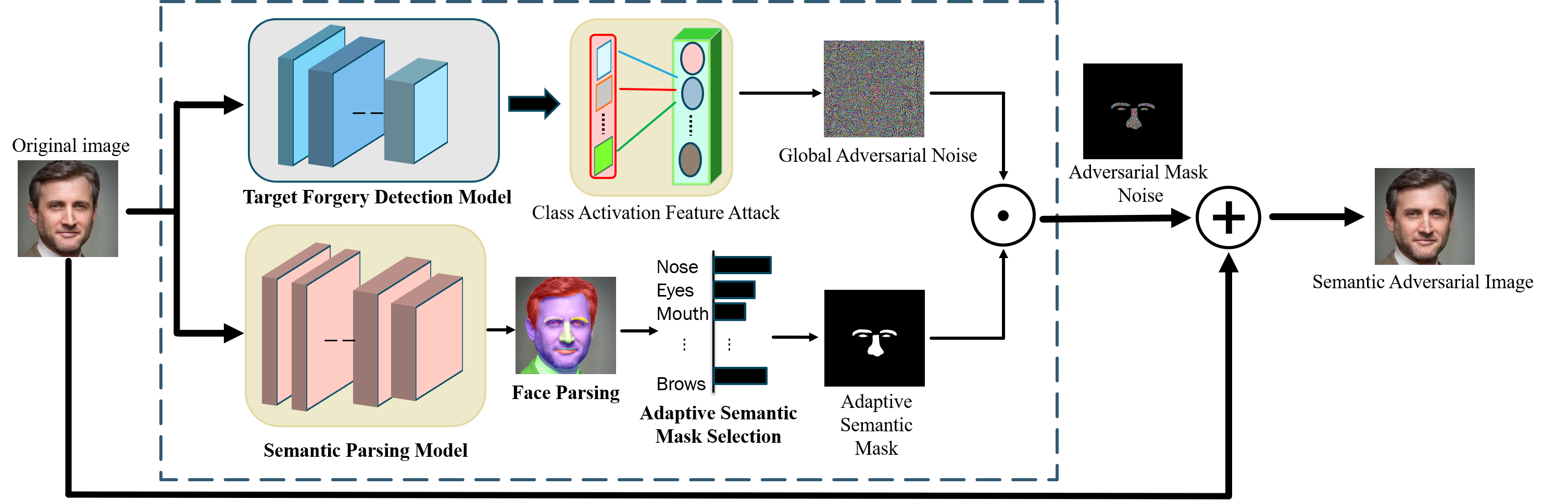}
    \caption{The framework of the proposed adversarial semantic mask attack for face forgery detection method.}
\label{fig:framework}
\end{figure*}

\subsection{Adversarial Semantic Mask Attack Generation}

The interpretability of neural networks has always been a hot topic~\cite{zhang2021survey}, while deep convolutional neural networks are usually considered as a black-box model, which makes it difficult to understand its internal mechanism. 
Class Activation Mapping (CAM) is a feature visualization technique typically used to acquire images with a specific type of architecture Response heat maps for classification models. 
The class activation mapping responds to the model's attention to the input image, and visualization techniques based on class activation mapping play an important role in understanding the working mechanism of the model. 
Considering the possibility that interfering class activation features may directly affect the target model's output, the algorithm uses class-activated features to create adversarial examples inspired from \cite{zhou2021removing}.
Given a natural sample $ x $, the algorithm first initializes the adversarial sample $ x_0 $ as $ x $. Inputting $ x $ and $ x_0 $ into the pre-trained forgery detection model yields the class-activated features $ \Phi_x $ and $ \Phi_{x^\prime} $ for both, and then computes their gradients using the class-activated feature distance $ \Delta(x, x^\prime) $:
\begin{equation}
\label{eq1}
g_t=\nabla x\Delta(x,x^\prime),
\end{equation}
where $ \Delta(\cdot) $ is the characteristic distance measure function for $ x^\prime_t $ is updated with the gradient to obtain$ x^\prime_{t+1} $ :
\begin{equation}
\label{eq2}
     x^\prime_{t+1}=x^\prime_t+\alpha\cdot sign(g_t).
\end{equation}

Here $ \alpha $ is the attack step size, which will be obtained by constraining the perturbation:
\begin{equation}
\label{eq3}
      x^\prime_{t+1}=clip(x^\prime_{t+1},x-\epsilon,x+\epsilon).
\end{equation}

The class activation mapping map $ CAM_c(w, h) $ of the input samples is then obtained to locate the fake attention, and the face is segmented using the face analysis module to segment the face. 
The pixel averages of the activation maps are computed for different semantic regions as the forgery correlation scores and sorting, selecting labels to obtain semantic masks, and restricting the generated adversarial noise to the semantic mask region. 
By iteratively performing such an update process, the algorithm can achieve maximizing $ \Delta(x, x^\prime) $ to obtain the final confrontation sample.
The final adversarial attack mask is acquired by integrating the selected face regions mask.
The training process and more algorithm details are shown in Algorithm~\ref{alg:algorithm}.

\subsection{Adaptive Semantic Mask Selection}
In order to select suitable face semantic regions to generate adversarial masks, we leverage the class activation values to choose semantic regions adaptively.
The face parsing network can perform semantic segmentation for the whole face image, classifying each pixel into a particular semantic label. 
We utilize the public algorithm to segment the face into multiple semantic categories, including left eye, right eye, left eyebrow, right eyebrow, nose, upper lip, lower lip, inside of mouth, face, hair, or background. 
The corresponding semantic features are extracted by selecting the labels corresponding to the facial regions.
Integrated with the class activation values of different regions, we select the most suitable face regions to generate an adversarial attack mask, which contains smaller and more important areas for face forgery detection tasks.

\section{Experiments}
In this section, we evaluated the proposed method on the dataset: Deepfake Detection Challenge dataset. 
We compared other state-of-the-art methods and the experimental results prove that our method achieved satisfactory performance in the image attack task.
Then we investigate the effect of different parameters on the recognition performance.
Finally, we conduct an ablation study to evaluate the effectiveness of the proposed ASMA.
\subsection{Databases}
DFDC dataset contains 472GB of data, including 119,197 face videos, of which 100,000 are fake face videos and 19,197 are videos taken by real people with more lifelike content. 
Of these, 100,000 videos are fake face videos and 19,197 videos are real videos with more lifelike content. 
The fake face videos were generated using a variety of face generation techniques, including DeepFakes, face2face, and other face-faking and expression editing algorithms, as well as unlearned methods to make the dataset contain as many fake face videos as possible. 
Each video in the dataset has a duration of 10 seconds, a frame rate ranging from 15 to 30 fps, and a video resolution ranging from 320×240 to 3840×2160, which makes the DFDC dataset more interesting than other datasets. 
The DFDC dataset has the largest size and the richest number of fake faces compared to other datasets.

\begin{algorithm}[!t]
    \caption{The detailed training process of ASMA.}
    \label{alg:algorithm}
    \textbf{Input}: A pretrained face forgery detection model $ \mathcal{P} $, input original face image $ x $. \\
    \textbf{Parameter}: number of iterations $ T $ and attack step size $ \alpha $. \\
    \textbf{Output}: The adversarial example $ \bar{x}$. \\  
\begin{algorithmic}[1] 
\STATE $ \bar{x}_0 \leftarrow x $.
\FOR{$ t $ = 0 to $ T $ - 1} 
\STATE Forward $ x $ and $ \bar{x}_t $ to $ \mathcal{P} $, and obtain class activation features $ \Theta_x $ and $ \Theta_{\bar{x}_t} $.
\STATE Compute the feature distance.
    \begin{center}
        $\Delta(x, \bar{x}_t)=\delta(\Phi_x,\Phi_{\bar{x}_t})$
    \end{center}
\STATE Compute gradients with the Eq.~(\ref{eq1}).
\STATE Update the adversarial example $ \bar{x}_t $ with the Eq.~(\ref{eq2}).
\STATE Project $ \bar{x}_{t+1} $ to the vicinity of $ x $ with the Eq.~(\ref{eq3}).
\ENDFOR
\STATE Compute global noise:
    \begin{center}
        ${x_g}=x_T-x$
    \end{center}
\STATE Generate semantic mask $ x_m $ by the adaptive semantic mask selection strategy.
\STATE Update the adversarial example $ \bar{x}_T $:
    \begin{center}
        $\bar{x}_T=x+{x_g} \odot x_m$
    \end{center}
\STATE \textbf{return} {$ \bar{x}_T $}.
\end{algorithmic}
\end{algorithm}

\subsection{Experimental Settings}

\begin{table*}[t]
    \caption{Evaluation of ASMA and other adversarial attack algorithms.}
    \label{tabel:semi}
    \centering
    \begin{tabular}{c|ccccc}
        \toprule
        Model & Method & \makecell{Xception\\ASR(\%)} & \makecell{ResNet-50\\ASR(\%)} & \makecell{EfficientNet-b0\\ASR(\%)} & \makecell{EfficientNet-b4\\ASR(\%)} \\\hline
        \multirow{7}{*}{XceptionNet}
        & FGSM\cite{goodfellow2014explaining}  & 42.04 & 13.19 & 2.85 & 9.70 \\
        & BIM\cite{kurakin2018adversarial}   & 85.20 & 17.92 & 5.94 & 14.63 \\
        & PGD\cite{madry2017towards}   & 85.23 & 18.47 & 6.14 & 15.88 \\
        & C\&W\cite{carlini2017towards}  & 68.39 & 11.10 & 1.25 & 0.05 \\
        & DeepFool\cite{moosavi2016deepfool}  & 77.62 & 8.30 & 0.42 & 0.18 \\
        & TRM\cite{Liu_2023_ICCV} & 81.19 & 16.25 & 7.45 & 19.16 \\
        & \textbf{ASMA}  & \textbf{85.33} & \textbf{31.68} & \textbf{11.46} & \textbf{23.02} \\
        \hline
        \multirow{7}{*}{ResNet-50}
        & FGSM\cite{goodfellow2014explaining}  & 9.35 & 74.84 & 10.04 & 14.91 \\
        & BIM\cite{kurakin2018adversarial}   & 16.44 & 75.49 & 10.96 & 17.66 \\
        & PGD \cite{madry2017towards}  & 16.66 & 75.54 & 11.94 & 17.72 \\
        & C\&W \cite{carlini2017towards} & 3.03 & 58.22 & 0.15 & 0.31 \\
        & DeepFool\cite{moosavi2016deepfool}    & 1.67 & 75.15 & 0.18 & 0.77 \\
        & TRM\cite{Liu_2023_ICCV}  & 15.76 & 66.24 & 19.63 & 22.95 \\
        & \textbf{ASMA} & \textbf{33.72} & \textbf{75.57} & \textbf{24.67} & \textbf{27.17} \\
        \hline
        \multirow{7}{*}{EfficientNet-b0} 
        & FGSM\cite{goodfellow2014explaining} & 16.86 & 28.23 & 10.04 & 25.69 \\
        & BIM\cite{kurakin2018adversarial} & 27.01 & 29.41 & 26.87 & 28.17 \\
        & PGD\cite{madry2017towards} & 27.70 & 31.26 & 26.91 & 28.20 \\
        & C\&W \cite{carlini2017towards}& 5.98 & 3.49 & 1.01 & 0.09 \\
        & DeepFool\cite{moosavi2016deepfool} & 4.03 & 3.45 & 1.45 & 0.16 \\
        & TRM\cite{Liu_2023_ICCV}  & 30.18 & 37.48 & 26.33 & 52.59 \\
        & \textbf{ASMA} & \textbf{34.90} & \textbf{58.05} & \textbf{43.05} & \textbf{62.25} \\
        \hline
        \multirow{7}{*}{EfficientNet-b4} 
        & FGSM\cite{goodfellow2014explaining} & 14.08 & 4.31 & 18.12 & 39.31 \\
        & BIM\cite{kurakin2018adversarial} & 21.18 & 5.49 & 20.91 & 47.78 \\
        & PGD\cite{madry2017towards} & 21.14 & 6.94 & 21.02 & 48.44 \\
        & C\&W \cite{carlini2017towards}& 9.92 & 1.53 & 0.21 & 16.37 \\
        & DeepFool\cite{moosavi2016deepfool} & 12.62 & 0.82 & 0.18 & 16.06 \\
        & TRM\cite{Liu_2023_ICCV}  & 25.95 & 27.56 & 28.33 & 52.54 \\
        & \textbf{ASMA} & \textbf{27.46} & \textbf{29.44} & \textbf{28.54} & \textbf{65.93} \\
        \bottomrule
    \end{tabular}
\end{table*}
\textbf{Implementation details}
During the experiments, 4000 videos in the dataset are randomly selected, and each video contains 300 frames. 
The algorithm in this chapter adopts MobileNet\_SSD as the face detection model, extracts and generates aligned face images frame by frame, and if more than one face is detected in a frame, only the largest face is extracted. 
To balance the influence of positive and negative sample imbalance, the algorithm randomly extracts 15 frames for fake face videos and 75 frames for real videos during training. 
Then the cropped faces are preprocessed, and all aligned face images are scaled to 320 × 320 and saved. 
To evaluate the adversarial performance of the algorithm, the experiment uses the above method to extract 1000 images of fake faces, which are used as inputs to the network to generate adversarial samples. 
The training iterations of the forgery detection model are 50 rounds, and the algorithm adopts the Adam optimizer, with the network hyper-parameters set to $ \beta_1 $= 0.9, $ \beta_2 $=0.999, and the learning rate $ l_r $=0.01. During the generation of the confrontation samples, the perturbation size $ \epsilon $= 0.20, the number of iterations $ T $ = 20, and the step size $ \alpha $ = 0.015.\\
\textbf{Target forgery detection models}
This paper uses some classical forgery detection models XceptionNet, ResNet50, EfficientNet-B0, and EfficientNet-B4 to study the ability to counter sample white-box and black-box attacks. 
This algorithm retrains these models on the DFDC dataset, where the inputs to all the forgery detection models are 320×320×3 images.

\subsection{Comparison Results}
This section compares this algorithm with representative adversarial attack algorithms~\cite{goodfellow2014explaining, kurakin2018adversarial, madry2017towards, carlini2017towards, moosavi2016deepfool, tang2023adversarial}. 
The algorithm generates adversarial examples based on the four classical models in the first column and transfers them to other networks for evaluation testing. 
As shown in Table~\ref{tabel:semi}, it can be seen that the face confrontation image generation algorithm based on semantic mask noise in this chapter has stronger attack performance compared to attack algorithms such as FGSM and PGD. 
The proposed ASMA has the highest white-box attack success rate compared to both FGSM and PGD methods, e.g., the success rate of the adversarial sample generated based on XceptionNet when attacking XceptionNet is 43.29\% higher than the FGSM method with a reduced attack area, 0.13\% higher than the BIM method, 0.1\% higher than the PGD method, 19.64\% higher than the C\&W method, 7.71\% higher than the DeepFool method, and 4.14\% higher than TRM method, which shows a very good performance of white-box attack. 
In comparison with other attack methods, the ASMA method produces adversarial examples with less variation in the attack success rate on different models, which indicates the good migratability of the adversarial examples.
Meanwhile, the adversarial examples generated by the method in this chapter on ResNet50 have an attack success rate 17.06\% higher than that of PGD when attacking XceptionNet, 12.73\% higher than that of PGD when attacking EfficientNet-B0, and 9.45\% higher than that of PGD when attacking EfficientNet-B4, which verifies the black-box migration performance of this algorithm. black-box migration of this algorithm. 
In addition, it can be seen that the success rate of the antagonistic samples generated on XceptionNet is lower when attacking EfficientNet-B4, and the adversarial examples generated based on EfficientNet-B4 are also difficult to migrate to XceptionNet due to the obvious structural difference between EfficientNet-B4 and XceptionNet, which is because the two networks have different structure. 
This is because of the obvious structural difference between EfficientNet-B4 and XceptionNet, and the adversarial attacks between the two networks have limited migration to each other. 
The experimental results show that adding a small amount of adversarial perturbation to the critical region of the image forgery has significantly improved the success rate of adversarial perturbation for the XceptionNet, ResNet50, EfficientNet-B0, and EfficientNet-B4 models. 

\subsection{Algorithm Analysis}

Table \ref{tabel:selc} shows experimental results after adding adversarial noise to different attribute regions of the face during the generation of adversarial samples, and the experiments reflect the impact of the perturbation region selection on the success rate of the attack and the visual quality. 
Since face forgery mainly tampers with the five senses of the face, the main attention of the forgery detection model is also focused on these regions, and the facial skin, eyes, nose, eyebrows, and hair are selected as the perturbation regions for the experiments. 
In our experiments, we reflect the model's prioritization of these regions by calculating the number of pixels in a particular region as a proportion of the overall pixels.
From the result of the experiment, we combine the success rate of the attack after adding noise to the facial regions, the visual effect, and the importance of the main facial recognition regions of the model, and finally choose the eyes, nose, and eyebrows as the joint attack regions.
As shown in the results, it can be observed that the success rate of the attack on skin and eyes is the highest among the perturbation selections of a single region, but due to the large area of the skin, the quality of the image is reduced after the attack, which makes the overall image less stealthy. 
As shown in the experimental results, it can be found that a better attack effect can be achieved by using a combination of multiple regions to attack.
The proposed algorithm selects a combination of selected regions for training, which is combined with the effect of the attack on the image quality, resulting in generating the most aggressive adversarial examples while maintaining a high degree of concealment. 

\begin{table}[!t]
    \caption{Quality analysis of adversarial images generated by selecting different features.}
    \centering
    \label{tabel:selc} 
    \begin{tabular}{c|c|c|c|c|c}
        \toprule
        Feature & MSE & MAE & PSNR & SSIM & Value Rate\\\hline
        Skin & 0.0725 & 0.0354 & 82.7299 & 0.8797 & 0.031\\
        Nose & 0.0692 & 0.0022 & 74.8923 & 0.9249 & 0.016\\
        Eye & 0.0651 & 0.0006 & 70.5987 & 0.9372 & 0.004\\
        Brow & 0.0636 & 0.0008 & 78.9069 & 0.9852 & 0.009\\
        Hair & 0.0527 & 0.0526 & 81.4216 & 0.9349 & 0.001\\
        Eye+Nose+Brow & 0.0757 & 0.0075 & 63.0589 & 0.8939 & 0.029\\
        \bottomrule
    \end{tabular}
\end{table}
\subsection{Parameter Analysis}
To evaluate the effect of the perturbation threshold size $ \epsilon $ on the model performance, the experiments in this section generate adversarial examples on XceptionNet, set 6 groups of different sizes of perturbation, the perturbation varies from 0 to 0.25, and the increase is 0.05 each time. 
The experiments test the success rate of the generated confrontation images of the face on the XceptionNet forgery detection model and count MSE, MAE, PSNR, SSIM, and other metrics to evaluate the algorithm attack performance and the quality of the generated images. 
From Table~\ref{tabel:eval}, it can be found that the larger the perturbation, the higher its attack success rate, but the worse the image quality index. 
When the perturbation is too large (greater than 0.15), as the perturbation increases, the gain in the attack success rate is not obvious but causes a rapid decline in the image quality metrics.  
Based on the above experimental results, the hyperparameter of the perturbation $ \epsilon $ is set to 0.15 in the adversarial attack module of the algorithm in this chapter for training and testing. 
\begin{table}[!t]
    \caption{Evaluation of different perturbation sizes in terms of attack success rate and visual quality.}
    \centering 
    \label{tabel:eval}
    \begin{tabular}{c|c|c|c|c|c}
        \toprule
        Perturbation & ASR(\%) & MSE & MAE & PSNR & SSIM \\
        \hline
        0.10 & 35.64 & 0.0118 & 0.2223 & 64.1251 & 0.9998 \\
        0.15 & 69.38 & 0.0387 & 0.1088 & 63.8271 & 0.9985 \\
        0.20 & 85.33 & 0.0757 & 0.0075 & 63.0589 & 0.9939 \\
        0.25 & 91.77 & 0.1254 & 0.0031 & 60.8061 & 0.9878 \\
        0.30 & 95.28 & 0.2181 & 0.0019 & 58.5085 & 0.9741 \\
        \bottomrule
    \end{tabular}
\end{table}

\subsection{Quantitative analysis}
\begin{figure}[h]
    \centering
    \includegraphics[width=0.48\textwidth]{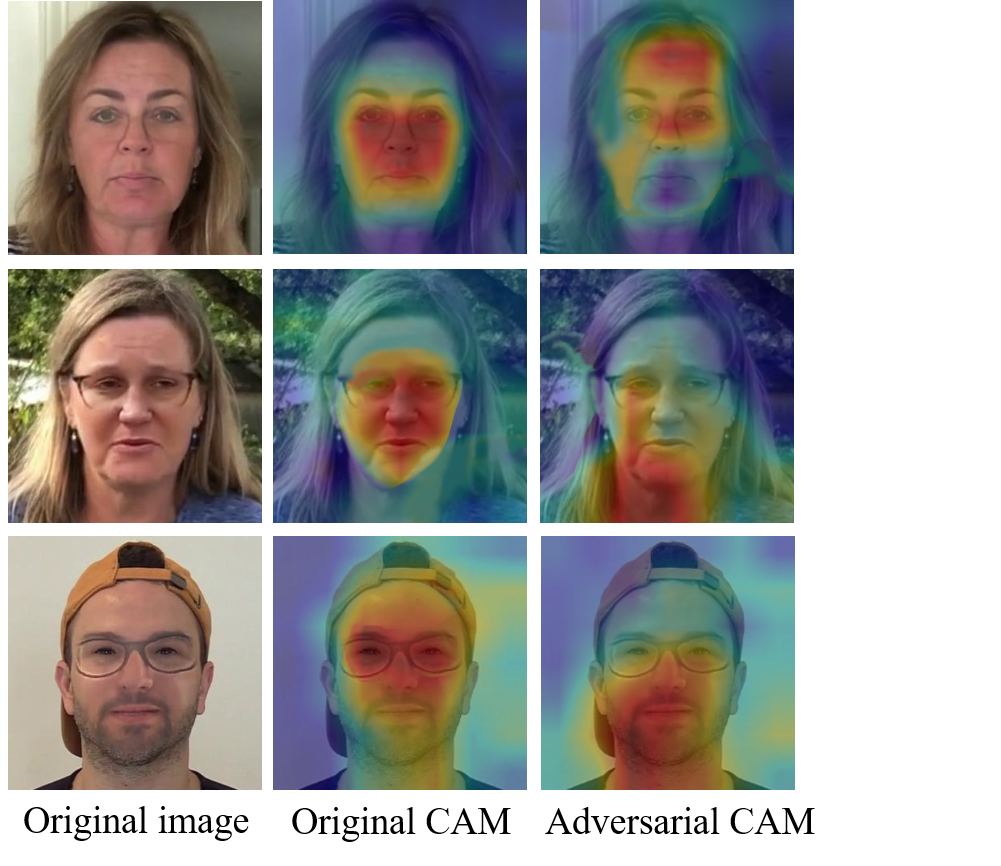}
    \caption{
        Visualization comparison of feature activation area changes before and after adversarial attacks on real faces.
    } \label{true}
\end{figure}

This section compares the quality of images generated by ASMA and other adversarial algorithms through qualitative and quantitative analysis. 
An important aspect of the quality requirements of adversarial images is concealability and different adversarial algorithms have different ways of generating images and generate images with inconsistent strength of concealment. 
Qualitative analysis experiments observe the difference between the generated adversarial image and the source image, combined with the added adversarial noise, to judge the adversarial generation algorithm. 
To better reflect the quality of the generated image, the commonly used image quality assessment indicators are used to calculate the difference from the original image. 
Quantitative experiments were compared by MSE, MAE, PSNR, and SSIM quality assessment metrics. 
For a better comparison of the results, we use Xception to generate the adversarial images, and for ASMA, set the maximum perturbation to 0.15 and the size of each iteration to 0.05.

\subsection{Qualitative analysis} 

\begin{figure}[h]
    \centering
    \includegraphics[width=0.48\textwidth]{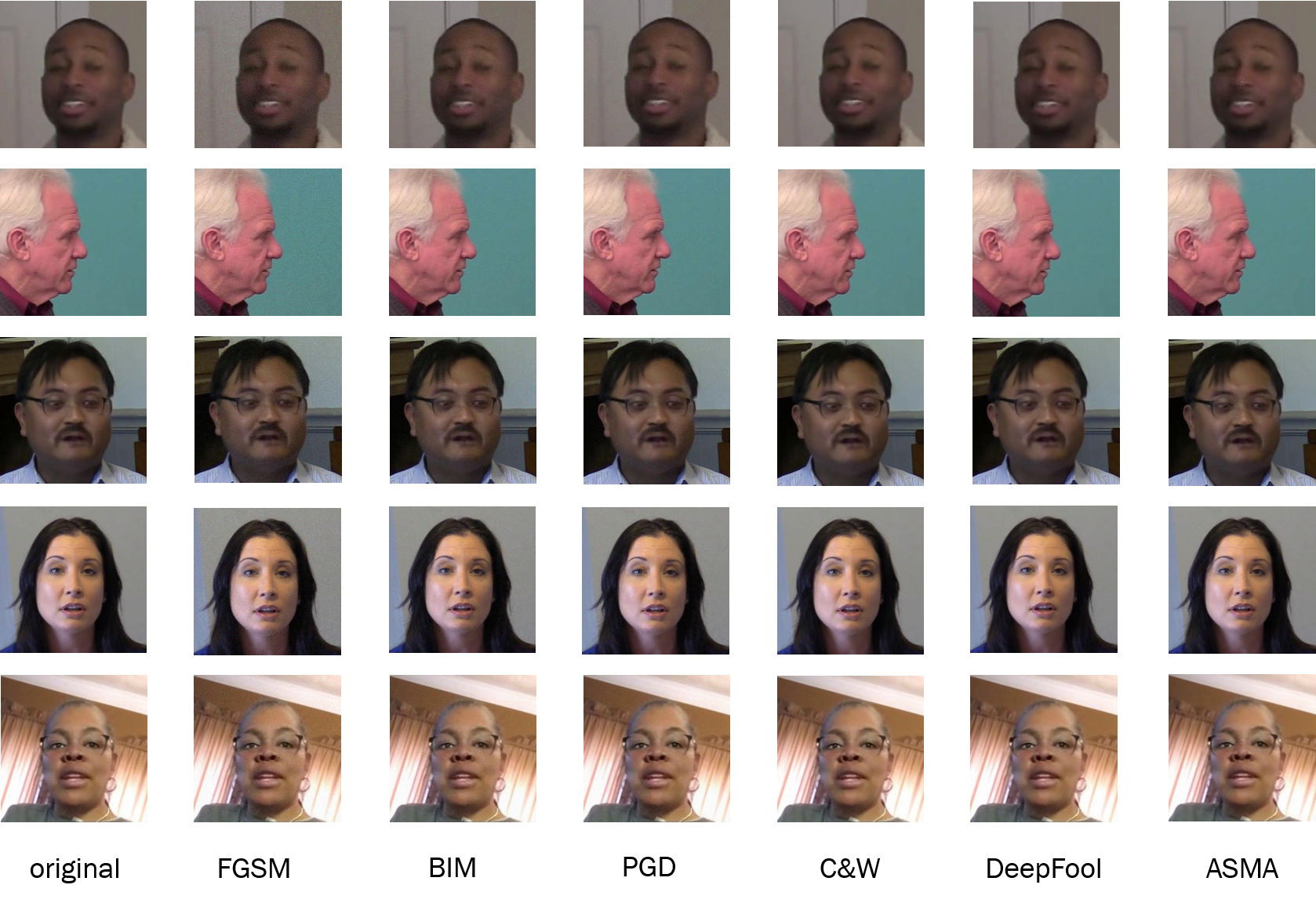}
    \caption{
     Comparison of images generated by different adversarial generation algorithms.
    } \label{comparison}
\end{figure}

The first column of Figure~\ref{comparison} is the original human face, and the rest is the adversarial face generated by FGSM, BIM, PGD, C\&W, DeepFool, and ASMA. 
In the attack process of ASMA, the nose, left and right eyes, and left and right eyebrow areas of the face are selected to add adversarial perturbation. 
Since these areas are forged key areas of face images and have more complex texture features, adding appropriate adversarial perturbations to this area has better attack performance, and adversarial perturbations are more difficult to perceive by the human eye. 
In this section, the masks of the left and right eyes and the left and right eyebrow regions are selected, and the semantic mask is multiplied by the adversarial noise so that the adversarial samples generated in each iteration only retain the adversarial perturbation of the mask region. 
Due to the decrease in the disturbance area, the attack performance may be reduced. To take into account the attack performance of the algorithm, the algorithm sets a reasonable threshold to constrain the size of the disturbance. 
As can be seen from Figure~\ref{comparison}, the algorithm in this chapter restricts the adversarial perturbation to the local semantic mask region, the image background and the face have no traces of perturbation, and the range of the adversarial perturbation region is significantly reduced compared to the perturbation of FGSM and PGD. 
The global adversarial perturbation generated by the FGSM and PGD algorithms adds too much adversarial noise in the background regions, which have simple image texture with a single color, making the generated adversarial samples with obvious adversarial texture, which can be easily detected by the human eye. 
After a qualitative comparison of the results generated from the confrontation samples, it can be seen that the confrontation generated by the algorithms in this chapter faces are more visually covert.

As shown in Table~\ref{tabel:quak}, we find that ASMA has significantly lower MSE, and MAE metrics than the rest of the methods through the addition of local mask noise. 
In addition, the images generated by the ASMA are more similar to the original images. 
It is worth noting that the perturbations generated by the algorithms such as FGSM, BIM, PGD, etc. are global, while the algorithm in this chapter generates the adversarial noise only in the semantic mask region, so the algorithm in this chapter is completely ahead of other algorithms in terms of the evaluation indexes of image quality. 
In addition, for the fairness of the comparison, the experiment compares the global adversarial faces generated by the algorithm in this chapter with other methods. 
ASMA(w/o mask) in Table~\ref{tabel:quak} represents that the algorithm does not use a semantic mask to constrain the noisy region, and iteratively generates the adversarial samples directly on the whole face image. 
Analysing the experimental results shows that the algorithm in this chapter has higher visual quality than existing methods even without constraining the perturbed regions. 
The experimental results show that the algorithm in this chapter can effectively improve the similarity between the face confrontation image and the original image with stronger concealment.

\begin{table}[t]
    \caption{Evaluation of images generated by different adversarial generation algorithms.}
    \label{tabel:quak}
    \centering
    \begin{tabular}{c|c|c|c|c}
        \toprule
        Method & MSE & MAE & PSNR & SSIM \\
        \hline
        FGSM & 54.0564 & 0.0879 & 30.0652 & 0.6631 \\
        BIM & 24.7829 & 0.0861 & 34.1893 & 0.8249 \\
        PGD & 26.7434 & 0.0645 & 34.8586 & 0.8132 \\
        C\&W & 0.5005 & 0.0254 & 51.1374 & 0.9795 \\
        DeepFool & 0.3723 & 0.0167 & 59.5134 & 0.9897 \\
        ASMA(w/o mask) & 63.6884 & 0.0859 & 33.0589 & 0.8839 \\
        \textbf{ASMA} & \textbf{0.0757} & \textbf{0.0075} & \textbf{63.0589} & \textbf{0.9939} \\
        \bottomrule
    \end{tabular}
\end{table}

\subsection{Visualization Analysis}
In Figure~\ref{true}, we can find that the detection model focuses on specific regions of the face on real faces. 
After the ASMA attack, the model would be misled during the face forgery detection process, it prefers to focus on other areas that are not attacked, so that the attention of the key areas is distracted, thus ensuring the invisibility of the adversarial examples. 
It can be seen from the experimental results that the model's attention to the image forgery region changes before and after the attack. 
In addition, it can be found from the original face CAM that the model's attention to the original fake face mainly focuses on the facial features of the face, especially the eye eyebrows. 
This is because the face forgery generation algorithm mainly tampers with the facial features, making the prediction of the model more sensitive to these areas. 
Therefore, adding subtle adversarial perturbations to these regions can shift the model's attention and mislead the model's prediction results. 
The comparison results of the class activation map also prove that the attack method based on class activation features is effective against the forgery detection model. 

\section{Conclusions}
In this paper, we propose an adversarial semantic mask attack framework (ASMA) for face forgery detection tasks.
The proposed method designs a novel adversarial semantic mask generation pipeline attacking face forgery detection, which aims to constrain generated perturbations in local semantic regions for good stealthiness.
To further improve the stealthiness and transferability performance, we design the adaptive semantic mask selection strategy, which leverages the class activation mapping to select more suitable adversarial semantic mask regions.
Experiments on public large-scale DFDC datasets illustrate the superior performance of the proposed ASMA.
In the future, we will evaluate the proposed method on more complex real scenarios to adapt to the needs of the real world.

\bibliographystyle{IEEEtran}
\bibliography{ref}
\end{document}